# Backbone is All You Need: Assessing Vulnerabilities of Frozen Foundation Models in Synthetic Image Forensics

Chiara Musso
University of Trento
Trento, Italy
chiara.musso@unitn.it

Joy Battocchio
University of Trento
Trento, Italy
joy.battocchio@unitn.it

Andrea Montibeller
University of Trento
Trento, Italy
andrea.montibeller@unitn.it

Giulia Boato
University of Trento
Trento, Italy
giulia.boato@unitn.it

## Abstract

As AI-generated synthetic images become increasingly realistic, Vision Transformers (ViTs) have emerged as a cornerstone of modern deepfake detection. However, the prevailing reliance on frozen, pre-trained backbones introduces a subtle yet critical vulnerability. In this work, we present the Surrogate Iterative Adversarial Attack (SIAA), a gray-box attack that exploits knowledge of the detector's ViT backbone alone and operates entirely within the target detector's feature space to craft highly effective adversarial examples. Through our experiments, involving multiple ViT-based detectors and diverse gray-box scenarios, including few-shot learning, complete training misalignment and attack transferability tests, we demonstrate that this vulnerability consistently yields high attack success rates, often approaching white-box performance. By doing so, we reveal that backbone knowledge alone is sufficient to undermine detector reliability, highlighting the urgent need for more resilient defenses in adversarial multimedia forensics.

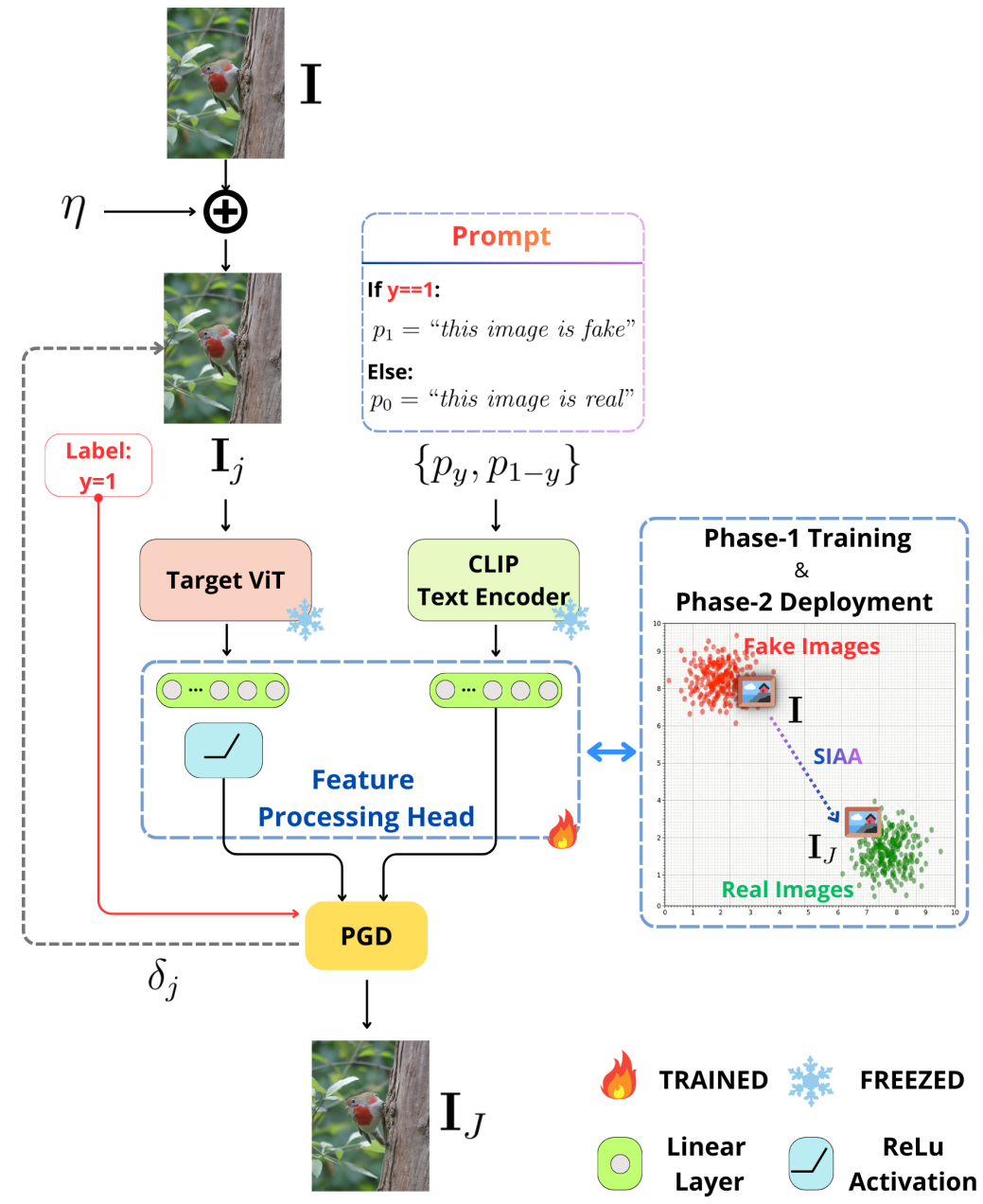

**Figure 1: Overview of the Surrogate Iterative Adversarial Attack (SIAA). In Phase 1, a surrogate Feature-Processing (FP)-head is trained to align frozen ViT features with CLIP text embeddings. In Phase 2, a Projected Gradient Descent (PGD) attack is used to optimize a perturbation $\delta_j$ by iteratively shifting the initial adversarial image $\mathbf{I}_j$ from the ground-truth cluster toward the target class. This process yields a final adversarial image $\mathbf{I}_J$ that bypasses the unknown classification head by exploiting learned feature distributions.**

## 1 Introduction

The rapid advancement of AI-based generative models [2] has enabled the creation of increasingly realistic synthetic images [15].

While these technologies facilitate creative applications, their potential for misuse (including identity theft, disinformation, and the production of child sexual abuse material) poses a significant threat to public trust and digital media integrity.

In response, the multimedia forensics community has focused on developing reliable AI-based synthetic image detectors [2]. Early approaches utilized convolutional neural networks (CNN) [8, 21] to capture low-level forensic artifacts to distinguish generated images from real ones. However, these methods often exhibit limited generalization across unseen generator classes, as their performance remains largely dependent on aligned training and testing conditions.

During the last two years, generalization challenges have been addressed by adopting large pre-trained backbones based on Vision Transformers [9, 16, 28]. In this paradigm, the ViT backbone is typically frozen, and a shallow classification head is trained to discriminate between real and synthetic images by exploiting high-level semantic features and mid-frequency artifacts. While joint fine-tuning can yield performance gains [7], it has been demonstrated that employing a frozen ViT backbone with a shallow classification head offers several computational advantages, including reduced hardware requirements during training and enhanced few-shot learning capabilities [5, 9].

Despite the high accuracy of ViT-based detectors, their adversarial vulnerability has received limited attention [4, 11, 13]. Recent studies [4, 11] have focused on benchmarking these detectors in terms of white-box robustness and transferability, revealing limited transferability between detectors with different architectures (such as CNN-based and ViT-based models) as well as comparable white-box robustness across both types. In contrast, vulnerabilities associated with gray-box attacks that specifically target frozen and pre-trained ViT backbones have been largely overlooked, as most existing attacks focus on other types of weaknesses, including frequency-domain artifacts [14] or general architectural weaknesses [17].

Nevertheless, the susceptibility of pre-trained ViT backbones to gray-box exploitation has already been demonstrated in the broader field of general-purpose image recognition. Arakelyan et al. [3] demonstrated this vulnerability by proposing a gray-box attack based on Projected Gradient Descent (PGD) [20], where the divergence between clean and adversarial feature representations is maximized by minimizing their cosine similarity within the target ViT backbone, without requiring access to the classification head or training data. A related no-box approach is explored by Zhang et al. [29], who achieved high Attack Success Rates (ASR) in object recognition tasks by relying on a CLIP [25] ViT backbone as a universal surrogate. However, these methodologies have yet to be adapted or evaluated for the unique constraints of synthetic image detection, where the decision boundary distinguishes authenticity rather than semantic classes.

To bridge this gap in adversarial multimedia forensics, we propose a novel gray-box attack called Surrogate Iterative Adversarial Attack (SIAA), depicted in Fig. 1. Inspired by [3], this attack assumes knowledge only of the pre-trained ViT backbone (and its weights) used by a fake image detector, and employs a surrogate model consisting of the CLIP text encoder, the target ViT backbone (used by the detector under attack), and a Feature Processing (FP) head composed of two separate linear layers: one for text embeddings and one for vision embeddings.

The attack is implemented in two phases. In the first phase, the FP-Head is trained on a dataset of real and fake images to learn a mapping that aligns CLIP text encoder embeddings and target ViT vision embeddings into shared clusters according to the image labels and the associated textual prompts (i.e., $label = 1$ "this image is fake", and $label = 0$ "this image is real"). Subsequently, the feature representations learned by the classification head are used to optimize an adversarial perturbation, implemented in this work using PGD [20], with the goal of fooling the attacked ViT-based fake image detector.

We note that, in contrast to [11] and [13], where the authors focused on white-box and transferability attacks and used the detector's logits to optimize the adversarial perturbation, the proposed SIAA exploits feature representations and their corresponding distributions learned during phase one of the attack. This represents a fundamental distinction, as we assume knowledge only of the target backbone and not of the type or depth of the detector's classification head. Furthermore, we emphasize (and will demonstrate) that applying the proposed attack directly on the features extracted by the target ViT vision encoder, without the use of the FP-head as done in [3], produces sub-optimal perturbations and lower ASRs.

Our findings demonstrate that the proposed gray-box attack consistently achieves high ASR across different detectors and attacker knowledge assumptions, including variations in training datasets, data augmentation strategies, different classification heads, and dataset size mismatches, such as in few-shot learning scenarios. In several of these cases, SIAA attains misclassification rates comparable to those of white-box attacks, demonstrating that, even in adversarial multimedia forensics, mere knowledge of the detector ViT backbone introduces significant adversarial vulnerability and highlights the need for more robust defensive strategies. Finally, we also evaluate transferability between backbones obtaining results in line with the state of the art of white-box attacks [3, 11].

The remainder of the paper is organized as follows: Section 2 details the proposed SIAA attack and the considered gray-box scenarios; Section 3 presents the experimental evaluation and ablations; and Section 4 concludes the paper and discusses future research directions.

## 2 Surrogate Iterative Adversarial Attack

To evaluate the severity of ViT-based gray-box vulnerability in the context of fake image detection, in this section, we propose a novel gray-box attack called Surrogate Iterative Adversarial Attack (SIAA)[1]. Comprehensive details regarding the training process, deployment, and hyperparameter settings are provided in Sects. 2.1 and 2.2.

As illustrated in Fig. 1, given an input image $\mathbf{I} \in \mathbb{R}^{H \times W \times 3}$ and its label $y \in \{0, 1\}$, where $y = 0$ corresponds to the prompt $p_0$ =*"The image is real"* and $y = 1$ to $p_1$ =*"The image is fake"*, SIAA aims during $J$ iterations, to generate an adversarial perturbation to fool the target fake image detector. However, unlike standard adversarial attacks that require access to the target detector's logits, SIAA operates entirely in the feature space, exploiting only knowledge of the detector's pre-trained ViT backbone. To do so, SIAA initializes the process by adding a random uniformly distributed perturbation $\eta \sim \mathcal{U}([-\varepsilon, \varepsilon]^{H\times W\times 3})$ (with $\epsilon$ the maximum allowable perturbation budget to ensure that the adversarial modifications remain imperceptible to the human eye) to $\mathbf{I}$, such that at iteration $j = 0$, the input is $\mathbf{I}_{j=0} = \mathbf{I} + \eta$. Thus, the proposed attack leverages the CLIP [25] text encoder $G(\cdot)$ to transform input prompts into a contextualized latent representation of size $D_t$, $\mathbf{t} = G(\{p_y, p_{1-y}\}) \in \mathbb{R}^{2\times D_t}$ (with $y$ equal to the image label), and the target detector's ViT backbone $F(\cdot)$ to extract high-dimensional visual features $\mathbf{v} = F(\mathbf{I}_j) \in \mathbb{R}^{B\times D_v}$, where $B$ is the batch size and $D_v$ is the feature dimension. Both $G(\cdot)$ and $F(\cdot)$ are kept frozen throughout the process.

[1] Code is available at: https://github.com/MMLab-unitn/SIAA-IHMMSec26

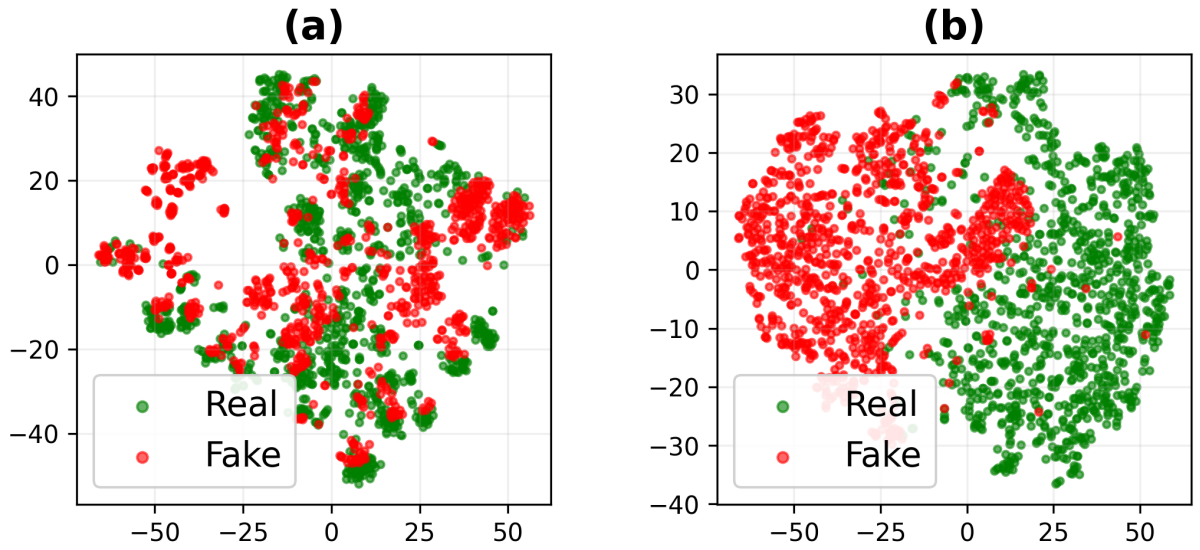

**Figure 2: t-Sne visualization of latent feature representations from a CLIP [9] ViT-based fake image detector: (a) output of the frozen CLIP ViT backbone, and (b) after the first trainable linear layer.**

Consequently, $\mathbf{t}$ and $\mathbf{v}$ are processed by a Feature-Processing (FP) head $H(\cdot)$ consisting of two specialized branches: $L_t : \mathbb{R}^{2\times D_t} \rightarrow \mathbb{R}^{2\times N}$, which processes the text features $\mathbf{t}$ to obtain $\mathbf{t}' = L_t(\mathbf{t})$ via a single linear layer; and $L_v : \mathbb{R}^{B\times D_v} \rightarrow \mathbb{R}^{B\times N}$, which processes the visual features $\mathbf{v}$ to obtain $\mathbf{v}' = L_v(\mathbf{v})$ via a linear layer and a ReLU activation. The objective of the FP-Head is to project $\mathbf{t}$ and $\mathbf{v}$ into a shared embedding space of dimension $N$, approximating the latent operations of the unknown attacked detector's classification head without requiring its specific architecture or weights. In fact, as shown in Fig. 2, it is the detector classification head that clusters the pretrained ViT features accordingly to the input image labels. Finally, $\mathbf{t}'$ and $\mathbf{v}'$ are used by PGD [20] to optimize, over $J$ iterations, the adversarial perturbation $\delta_j$ to be applied to $\mathbf{I}_j$ and attack the target detector.

For this task, we selected the PGD attack under an $\ell_\infty$-norm constraint , ensuring $\|\delta_{j=J}\|_\infty \leq \epsilon$. Nevertheless, as we assume only knowledge of the ViT backbone, we depart from the common practice of maximizing a cross-entropy loss on logits, and opt for a custom distance-based objective function $\mathcal{J}$ which operates in the surrogate feature space defined by the FP-Head:

$$\mathcal{J} = \|\mathbf{v}' - \mathbf{t}'_y\|_2 - \|\mathbf{v}' - \mathbf{t}'_{1-y}\|_2, \tag{1}$$

where $\mathbf{t}'_y$ is the processed textual embedding of the ground-truth class, and $\mathbf{t}'_{1-y}$ is the textual embedding of the adversarial target class. By maximizing $\mathcal{J}$ , the attack pushes the visual representation $\mathbf{v}'$ away from the ground-truth class text embedding and toward the adversarial target class embedding. We note that, the PGD attack optimization of (1) is the opposite of the approach adopted to train foundational ViT-based architectures [25].

Consequently, at each iteration $j \in \{1, \dots, J\}$, PGD updates the image as:

$$\mathbf{I}_{j+1} = \mathrm{Proj}_{\mathbf{I},\epsilon}\left(\mathbf{I}_j + \alpha \cdot \mathrm{sign}\left(\nabla_{\mathbf{I}}\mathcal{J}\right)\right), \tag{2}$$

where, the PGD perturbation at iteration $j$ is $\delta_j = \alpha \cdot \mathrm{sign}\left(\nabla_{\mathbf{I}}\mathcal{J}\right)$, $\alpha$ is the step size, and $\mathrm{Proj}_{\mathbf{I},\epsilon}$ denotes the projection onto the $\ell_\infty$ $\epsilon$-ball around $\mathbf{I}$ and the valid pixel range $[0, 1]$. The final PGD perturbation is $\delta_J$ and the resulting adversarial image is $\mathbf{I}_J$.

## 2.1 SIAA: Training and Deployment

As introduced in Sect. 1, SIAA operates in two distinct phases.

*Phase One: Training.* We train the FP-Head $H(\cdot)$ while keeping the target ViT backbone $F(\cdot)$ and the CLIP text encoder $G(\cdot)$ weights frozen. The goal is to align visual $\mathbf{v}'$ and textual $\mathbf{t}'$ features in a shared embedding space that approximates the decision boundaries used by the target classification head.

Given the ground-truth label $y$, the FP-Head is optimized via a multi-term contrastive custom loss:

$$\mathcal{L} = \beta(e)\left(\mathcal{L}_{\text{pull}} + \mathcal{L}_{\text{push}}\right) + \mathcal{L}_{\text{text}}, \tag{3}$$

where the individual loss terms are defined as:

$$\mathcal{L}_{\text{pull}} = \|\mathbf{v}' - \mathbf{t}'_y\|_2^2 \tag{4}$$

$$\mathcal{L}_{\text{push}} = \left[\max(0, m - \|\mathbf{v}' - \mathbf{t}'_{1-y}\|_2)\right]^2 \tag{5}$$

$$\mathcal{L}_{\text{text}} = \max(0, 2m - \|\mathbf{t}'_y - \mathbf{t}'_{1-y}\|_2). \tag{6}$$

Here, $\mathcal{L}_{\text{pull}}$ minimizes the distance between the visual features $\mathbf{v}'$ and the textual features of the corresponding ground-truth class $\mathbf{t}'_y$, while $\mathcal{L}_{\text{push}}$ enforces a minimum margin $m$ relative to the contrastive class, $\mathbf{t}'_{1-y}$. The quadratic formulation of these terms ensures a stronger penalty for embeddings far from their respective class representations, prioritizing the tight clustering of visual features around their respective textual embeddings.

Conversely, $\mathcal{L}_{\text{text}}$ ensures that $\mathbf{t}'_y$ and $\mathbf{t}'_{1-y}$ remain separated by a margin of $2m$ (i.e. the minimum distance required to prevent overlap between two clusters of radius $m$), thereby enhancing attack transferability. We avoided quadratic penalty for $\mathcal{L}_{\text{text}}$, to prevent $\mathcal{L}_{\text{text}}$ magnitude to outweighing $\mathcal{L}_{\text{pull}}$ and $\mathcal{L}_{\text{push}}$ during the optimization process.

Finally, to maintain stability during the training process, we utilize a weighting schedule $\beta(e) = \min(1, e/\lfloor E/2 \rfloor)$ that gradually increases the influence of the image-based loss terms over $E$ training epochs, with $e$ the current training epoch.

*Phase Two: Deployment.* Once the FP-Head is trained, for any input image $\mathbf{I}$, the proposed attack uses the feature representations of $\mathbf{v}$ and $\mathbf{t}$, and the relative space learned by the FP-Head during phase one, in combination with Eq. (1), to optimize the PGD adversarial perturbation over $J$ iterations. We reiterate that, unlike conventional PGD attacks, SIAA operates solely on feature vectors, assuming no knowledge of the target classification head's size or weights. Once the PGD optimization reaches $J$ iterations, the final adversarial image is $\mathbf{I}_J$.

We evaluate the perceptibility of the proposed attack on images from the Synthbuster Dataset [9], producing 1000 adversarial images with an average Structural Similarity Index Measure (SSIM) of 0.989 and Weighted Peak Signal-to-Noise Ratio (WPSNR) of 43.956, in line with accepted values presented in the literature [11].

## 2.2 Hyperparameters & Training Setup

Unless otherwise specified, we trained our attack using the dataset from [11], comprising real images from MS-COCO [18] and generated counterparts from latent diffusion [27]. We used 10k real and 10k fake images for training, and additional 2k of each class reserved for validation (no overlap).

The surrogate head is trained for $E = 20$ epochs using Adam ($lr = 5 \times 10^{-3}$), with margin $m = 1$. All embeddings are $\ell_2$-normalized prior to loss computation. Data augmentation implements geometric, photometric, and noise transforms. Geometric transforms include 90° rotations ($p = 1.0$), horizontal flips ($p = 0.5$), and random resized crops to $224 \times 224$. Feature invariance is promoted via color jitter ($p = 0.8$), Gaussian blur ($p = 0.5$), compression artifacts ($p = 0.5$), random grayscale, cutout, and additive noise (each $p = 0.2$). Images are normalized according to each backbone's requirements.

Training and testing batch sizes $B$ are set to 64 and 1, respectively. Embedding dimensions are defined as $D_t = 768$, $D_v = 1024$, with the FP-Head $H(\cdot)$ outputting $N = 1024$ dimensional features. The maximum PGD iterations $J = 14$ is consistent with [3] to ensure a fair comparison, and $\epsilon = 8/255$. Final model selection is based on peak validation performance. Experiments were conducted on a workstation with an NVIDIA GeForce RTX 3090 GPU, an Intel(R) Core(TM) i9-10940X CPU, and 256 GB RAM.

# 3 Experimental Results

In this section, we evaluate the proposed attack under different levels of gray-box knowledge and compare it against [3] and a variant thereof implemented for this study. Specifically, we first present results under optimal gray-box conditions, where both the adversarial attacks and the detectors are trained on the same dataset using the same data augmentation pipeline. Thus, through ablation studies, we evaluate the robustness of SIAA under more restrictive scenarios, including variations in the detector's classification-head architecture, training under few-shot conditions, and cases where the training dataset and data augmentations are unknown. Finally, we present transferability results, including scenarios where the target backbone is unknown to the attacker, thereby assessing the attack's performance in true black-box settings.

All evaluations are performed in terms of Attack Success Rate (ASR), computed exclusively on images correctly classified by the detector prior to the attack, and Area Under the Curve (AUC), measured under both clean and adversarial conditions. The selected detectors are based on ViT backbones including: CLIP [25], DINOv2 [7], and Swin [19]; and incorporate (unless specified differently) a two-layer classification head, following the implementation in [9].

In all experiments, and for testing purposes exclusively, we used the Synthbuster dataset [9], comprising 9,000 synthetic images generated by nine state-of-the-art generative models: DALL-E 2 [26], DALL-E 3 [6], Adobe Firefly [1], GLIDE [23], Midjourney v5 [22], and Stable Diffusion (v1.3, v1.4, v2.0, and XL) [27]; supplemented by 1,000 real images from the RAISE dataset [10].

| Attack | **CLIP** [25] | | | **Swin** [19] | | | **DINOv2** [7] | | |
|---|---|---|---|---|---|---|---|---|---|
| | AUC | | ASR | AUC | | ASR | AUC | | ASR |
| | Pre | Post | | Pre | Post | | Pre | Post | |
| PGD [20] | 0.88 | 0.00 | 1.00 | 0.79 | 0.00 | 1.00 | 0.83 | 0.32 | 0.62 |
| PGD-ViT [3] | | 0.54 | 0.50 | | 0.47 | 0.51 | | 0.63 | 0.28 |
| PGD-L | 0.88 | 0.23 | 0.83 | 0.79 | 0.21 | 0.84 | 0.83 | 0.5 | 0.49 |
| SIAA | | **0.00** | **1.00** | | **0.00** | **0.99** | | **0.49** | **0.56** |

**Table 1: Gray-box attack performance (post-attack AUC and ASR). White-box PGD [20] serves as a performance upper bound. Bold values indicate lowest AUC and best ASR.**

## 3.1 Gray-Box Attack Success Rate

In the experiments reported in Table 1, we compare, in terms of ASR and AUC, the proposed SIAA against other gray-box attacks targeting detectors based on CLIP [25], Swin [19], and DINOv2 [7] ViTs. As the first gray-box baseline, we selected the adversarial attack proposed in [3], which we refer to as PGD-ViT, since it optimizes PGD perturbations directly on the output features of the target ViT. In addition, we extend PGD-ViT to define a third baseline, called PGD-L (Linear). PGD-L train and adds a linear layer (identical to the one used in the ViT branch by the FP-Head) on top of the target ViT, and optimizes the adversarial perturbation in this projected feature space. Finally, we report the results of PGD [20] in a white-box setting, which serves as an empirical upper bound. All detectors and attack variants were trained on the same dataset using identical data augmentation protocols. Details on the implications of a total misalignment in terms of dataset and augmentation are discussed in Sect. 3.2.

As seen from Table 1, SIAA is significantly more successful than the other baselines, PGD-ViT and PGD-L, achieving ASR and post-attack AUC values comparable to those of PGD [20], and demonstrating how vulnerabilities introduced by the mere knowledge of the detector can be exploited in adversarial multimedia forensics.

As discussed in Sect. 2, the superior ASR achieved by SIAA stems from the FP-Head's fusion of textual and visual features, which facilitates the learning of a hyperspace where adversarial attacks are significantly more effective. In fact, while fake image detectors show substantial robustness to gray-box attacks like PGD-ViT (which targets sparse ViT output features), this resilience is immediately undermined when perturbations are optimized on the subsequent linear layer (PGD-L), and is further diminished by SIAA.

Despite SIAA is the best-performing attack against all detectors, against DINOv2 [7], both SIAA and PGD [20] never exceed an ASR of 0.62. This observed robustness can be attributed to the intrinsic stability of DINOv2's self-supervised feature space. Unlike CLIP [25], which prioritizes vision-language alignment, or Swin [19], which often relies on supervised textural shortcuts, DINOv2 employs combined image- and patch-level objectives that compel the backbone to learn dense, object-centric structural representations [24]. Furthermore, DINOv2 incorporates register tokens to mitigate high-norm artifacts in attention maps. Consequently, these mechanisms likely induce a smoother loss landscape, eliminating the sharp gradients typically exploited by attacks like PGD [20].

## 3.2 Ablations

In these ablation studies, we evaluate the ASR of SIAA in contexts of partial or total misalignment with the training configuration of the target detector. Specifically, we evaluate SIAA against the detector based on the CLIP ViT pre-trained backbone, which yielded the highest ASR in Table 1.

In Table 2, we report the results obtained when one or multiple of the following levels $l$ of gray-box knowledge are considered. Under $l_1$, we consider variable classification head depth (i.e., 1, 2, or 3 layers) implemented by the target fake image detector; $l_2$ evaluates the performance of SIAA when it has access to only 100 real and 100 fake images from the original training dataset. In contrast, $l_3$ assumes that the implemented data augmentation is unknown (in which case, we do not employ any data augmentation), and $l_4$ assumes that the training dataset used for SIAA is different from the one used by the target detector. Specifically, for $l_4$, we train SIAA on 10k fake images generated by Stable Diffusion 2.1 and 10k real images equally sampled from the FORLAB and FFHQ classes of the TrueFake Dataset [12]. To ensure parity with the real data, these fake images were randomly sampled to balance the image content across diverse categories (i.e., faces, landscapes, animals, and general scenes) [12].

Analyzing the results for $l_1$ in Table 2, we observe that, when SIAA is trained on the same dataset and the same data-augmentation as the target detector, the feature representation learned by the FP-Head is robust to variations of the detector classification-head, with ASR$\geq$ 0.981. This result is particularly relevant as it shows that the gray-box vulnerability presented in this paper cannot be easily mitigated by the architectural choices defined by $l_1$. Additionally, ASR$\geq$ 0.96 are achieved also when SIAA is trained under the conditions described by $l_2$ (on just 100 + 100 training samples) and $l_3$, where we assume zero knowledge of the data-augmentation used to train the target fake image detector. In details, SIAA results particularly effective under the few-shot learning conditions of $l_2$, allowing for the definition of a feature space that is sufficiently representative also under limited training resources. This remains true also when $l_2$ and $l_3$ are considered simultaneously.

Finally, we note a more prominent ASR degradation under $l_4$, characterized by a total reduction of 0.07 in post-attack AUC and an ASR drop of 0.15 when comparing these results to the baseline reported in Table 1. Additional, performance decay can be observed under the combination of $l_4$ and $l_3$, where post-attack AUC and ASR exhibit total decreases of 0.28 and 0.3, respectively.

These results are likely attributed to the greater misalignment in the feature space and feature representations stemming from $l_4$, and necessitate the exploration of more advanced solutions, including domain-invariant feature alignment and self-supervised adaptation techniques. Nevertheless, we remark that, even under conditions of total misalignment (i.e., $l_3+l_4$) with respect to the detector's training phase, SIAA still achieves an encouraging ASR of 0.7, effectively attacking more than 50% of the images correctly classified by the tested fake image detector.

## 3.3 Transferability

After the promising results obtained in Sect. 3.2, we evaluated the cross-model transferability of SIAA perturbations to detectors employing backbones distinct from those it was trained on. Specifically, as was done in Sect. 3.1, SIAA was optimized against "source" backbones based on CLIP, DINOv2, and Swin. Then, we evaluated the resulting adversarial samples in terms of ASR and AUC against "target" detectors utilizing different ViT backbones.

| Ablation Events | | | | Metrics | |
|---|---|---|---|---|---|
| $l_1$ | $l_2$ | $l_3$ | $l_4$ | AUC (Post) | ASR |
| 1 | ALL | ✗ | ✗ | 0.05 | 0.981 |
| 2 | ALL | ✗ | ✗ | 0.00 | 1.000 |
| 3 | ALL | ✗ | ✗ | 0.03 | 0.990 |
| 2 | ALL | ✓ | ✗ | 0.02 | 0.985 |
| 2 | 100+100 | ✗ | ✗ | 0.02 | 0.985 |
| 2 | 100+100 | ✓ | ✗ | 0.01 | 0.960 |
| 2 | 10k+10k | ✗ | ✓ | 0.07 | 0.850 |
| 2 | 10k+10k | ✓ | ✓ | 0.28 | 0.700 |

**Table 2: Attack Success Rate (ASR) of SIAA targeting a CLIP-based ViT detector under varying levels of training misalignment. For conditions $l_3$ (unknown augmentation) and $l_4$ (misaligned dataset), symbols ✓and ✗indicate whether the event occurred. For $l_1$ and $l_2$, the classification head depth and dataset size are reported directly.**

The results, illustrated in Fig. 3, reveal a significant disparity in forensic vulnerability across models. First, we observe a promising attack transferability between Swin and CLIP, with ASR values of $\sim$ 0.56 and AUC values of $\sim$ 0.350. These results align with transferability trends reported in the literature [4, 11] for white-box attacks and suggest that SIAA effectively exploits a shared latent manifold between these backbones, likely due to similarities in their pre-training objectives.

On the other hand, DINOv2 exhibits substantial resistance to attack transferability. Specifically, when DINOv2 serves as a target, SIAA achieves an ASR of only 0.08–0.10, while the post-attack AUC hovers near 0.600–0.650. As discussed in Sect. 3.1, this increased robustness likely stems from DINOv2's architectural refinements and the use of register tokens to mitigate high-norm artifacts.

Interestingly, a different behavior is observed when DINOv2 is utilized as the source of the attack. In fact, despite its resistance as a target, perturbations optimized on the DINOv2 backbone maintain moderate transferability, achieving ASRs of 0.31 on CLIP and 0.28 on Swin. This indicates that while DINOv2 makes it difficult to optimize successful adversarial perturbations into its feature space, the forensic representations it learns are still sufficiently representative to craft attacks that can generalize to more vulnerable backbones.

# 4 Conclusions

In this paper, we introduced a novel gray-box attack called the Surrogate Iterative Adversarial Attack (SIAA), which exploits the predictable feature space of fake image detectors built on pre-trained Vision Transformer (ViT) backbones. Through our experiments, we showed that by using an FP-Head to align visual and textual embeddings, it is possible to craft highly effective adversarial attacks that operate entirely within the feature space of the detector, without requiring access to its classification head. Specifically, SIAA outperforms existing attacks in the literature, achieving ASRs comparable to white-box methods, even under extreme low-data regimes. Moreover, SIAA maintains ASRs $\geq$ 0.7 even when there is a misalignment between the training set and the data augmentation used by the target detector. While there is room for further optimization regarding ASR transferability to DINOv2, SIAA demonstrates promising cross-architecture performance (comparable to established white-box transferability benchmarks) when transferring between CLIP and Swin Transformer backbones, as well as from DINOv2 to both CLIP and Swin.

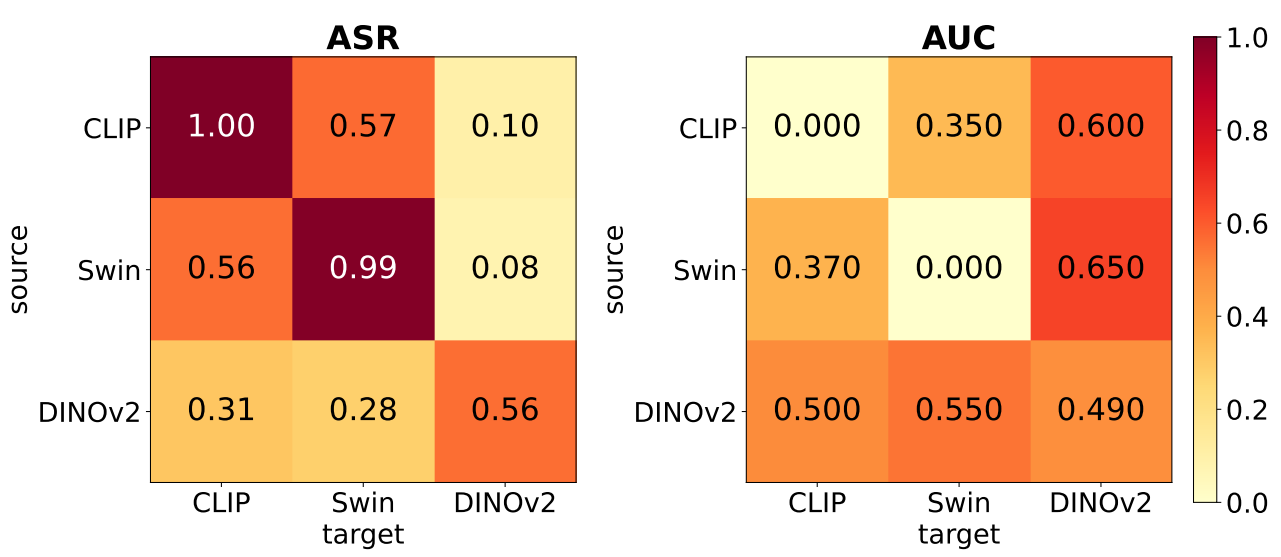

**Figure 3: Transferability of SIAA across different ViT backbones. The matrices report the ASR (left) and post-attack AUC (right) for perturbations optimized on a "source" backbone and evaluated against a "target" detector. Diagonal elements represent the gray-box scenario where source and target architectures are aligned.**

Building on the robust feature representations of DINOv2, future work will investigate whether its inherently smoother loss landscape can be exploited to design more effective adversarial perturbations. Specifically, we aim to bridge the performance gap observed when targeting models equipped with aggressive artifact mitigation. Furthermore, we plan to extend this gray-box framework to evaluate ViT-based detectors that have been fine-tuned in an end-to-end fashion. Finally, future work will explore the development of robust countermeasures to protect ViT-based fake image detectors against this class of gray-box attacks.

## Acknowledgments

This work was partially supported by the European Union under the Italian National Recovery and Resilience Plan (NRRP) of NextGenerationEU (PE00000014 - program "SERICS"), project SOS-AI.